\pdfoutput=1

\documentclass[11pt]{article}

\usepackage[final]{acl}
\usepackage{times}
\usepackage{latexsym}
\usepackage{multirow}
\usepackage{amsmath}
\usepackage{array}
\usepackage[T1]{fontenc}
\usepackage{subcaption} 
\usepackage[utf8]{inputenc}

\usepackage{microtype}

\usepackage{inconsolata}

\usepackage{graphicx}
\usepackage{amsfonts}  
\usepackage{amssymb} 
\title{Mitigating Biases in Language Models via Bias Unlearning}

\author{
 \textbf{Dianqing Liu\textsuperscript{1,2}},
 \textbf{Yi Liu\textsuperscript{2,}}\thanks{Corresponding author. The codes are available at https://github.com/a101269/BiasUnlearn.},
 \textbf{Guoqing Jin\textsuperscript{2}},
 \textbf{Zhendong Mao\textsuperscript{1}},
\\
 \textsuperscript{1}University of Science and 
Technology of China
\\
 \textsuperscript{2}State Key Laboratory of Communication Content Cognition, People’s Daily Online
\\
 \small{
 \texttt{ldqblcu@126.com},  \texttt{gavin1332@gmail.com}, 
  \texttt{jinguoqing@people.cn},  \texttt{zdmao@ustc.edu.cn}
 }
}

\begin{document}
\maketitle
\begin{abstract}
Many studies have shown various biases targeting different demographic groups in language models, amplifying discrimination and harming fairness. Recent parameter modification debiasing approaches significantly degrade core capabilities such as text coherence and task accuracy. And Prompt-based debiasing methods, only effective for predefined trigger words, fail to address deeply embedded stereotypical associations in model parameters. In this paper, we propose BiasUnlearn, a novel model debiasing framework which achieves targeted debiasing via dual-pathway unlearning mechanisms coordinating stereotype forgetting with anti-stereotype retention, while preventing bias polarity reversal through adversarial forget set and dynamic dataset swapping. We conducted extensive experiments with multiple language models across various evaluation benchmarks. The results show that BiasUnlearn outperforms existing methods in mitigating bias in language models while retaining language modeling capabilities. Further experiments reveal that debiasing weights are transferable across model variants, confirming that bias representations become entrenched during pre-training and persist through fine-tuning phases. 
\end{abstract}

\begin{figure}[ht]  
  \centering  

  \begin{subfigure}[b]{\linewidth}  
    \includegraphics[width=\linewidth]{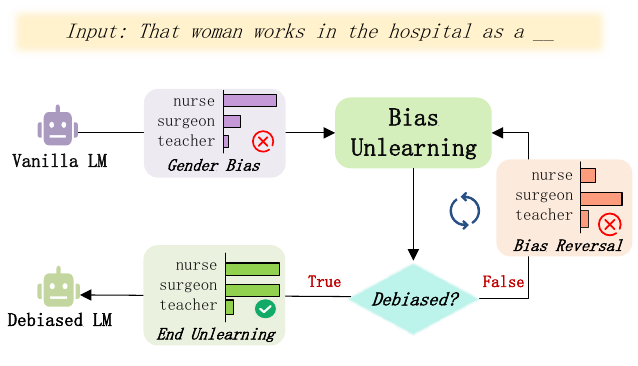} 
    \caption{The training process of BiasUnlean. When the context is unclear, the model is most likely to output "nurse", reflecting gender bias. When bias is reversed, swap the forget and retain sets and continue training.}
    \label{fig:sub_a}
  \end{subfigure}

  \vspace{1em}

  \begin{subfigure}[b]{\linewidth}  
    \includegraphics[width=\linewidth]{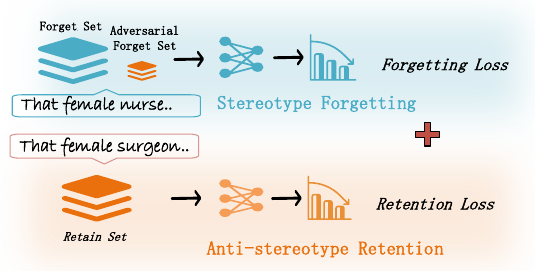} 
    \caption{BiasUnlearn incorporates dual-pathway unlearning mechanisms coordinating stereotype forgetting with
anti-stereotype retention, while preventing bias
polarity reversal through adversarial forget set.}
    \label{fig:sub_b}
  \end{subfigure}

  \caption{Demonstration of BiasUnlearn framework.}
  \label{fig:overall}
\end{figure}

\section{Introduction}

In recent years, large language models (LLMs) have been widely utilized in various fields. However, stereotypes in corpora constructed by human language inevitably affect these language models. Many studies have pointed out that there are significant social biases or stereotypes in large models, including gender bias, racial bias, and religious bias (\citealp{hofmann2024ai}; \citealp{kim2024lifetox}; \citealp{liu2023uncovering}; \citealp{hosseini2023empirical}; \citealp{esiobu2023robbie}). These biases are subtle and seemingly harmless on the surface, but they may have negative impacts on many applications. For example, translation software from companies like Google tends to translate gender-neutral pronouns in many languages into "he" in English without sufficient contextual \citep{zou2018ai}. When generating computer programs, LLMs can generate stereotypical content, which may cause displeasure to specific demographic groups. As reported by \citet{shrawgi2024uncovering}, stereotypes have a negative impact on the accuracy of model text classification, and reducing stereotype bias can effectively improve text classification performance \citep{shen2023words}. Consequently, the elimination of bias to ensure the fairness of model outputs, thereby preventing systematic discrimination within models, and enabling artificial intelligence technology to genuinely serve all of humanity is of great significance.

Debiasing involves balancing the output probabilities of stereotypical and non-stereotypical terms in LLMs when the context involves specific target groups. Many methods have been proposed to debias language models, such as re-pretraining on counterfactual data (\citealp{lu2020gender}; \citealp{zmigrod2019counterfactual}), debiasing with representation projection (\citealp{kaneko2021debiasing}; \citealp{ravfogel2020null}; \citealp{liang2020towards}), prompt-based methods (\citealp{schick2021self}; \citealp{furniturewala2024thinking}; \citealp{banerjee2024all}; \citealp{liprompting}), and Model-editing approach  \citep{xu2025biasedit}. However, these methods often fall short in effectively eliminating bias, preserving model capabilities, types of applicable models and ensuring efficient training and inference. For example, re-pre-training is effective but needs a great number of computing resources and time. Debiasing with representation projection remains limited in effectiveness due to the difficulty of obtaining high-quality bias representations. Prompt-based methods typically require multiple inferences per input, making them inefficient and limiting their ability to adapt to different bias types due to prompt limitations. The model-editing approach is efficient, but it comes at the expense of weakened language modeling capabilities and can have unintended side effects \cite{halevy2024flex}. In addition, re-pre-training can only be applied to pretrained models, while prompt-based methods can only be applied to instruction-fine-tuned models. 

In this paper, we propose BiasUnlearn,  an efficient and robust debiasing framework that mitigates biases in language models through targeted unlearning training. BiasUnlearn incorporates dual-pathway unlearning mechanisms that coordinate stereotype forgetting with anti-stereotype retention. Additionally, it safeguards against bias polarity reversal through adversarial forget set construction and dynamic dataset swapping. Experimental results demonstrate that these measures can effectively reduce social biases in language models while maximizing the retention of the model's capabilities. Further experiments reveal that the debiasing weights trained on the base models maintain robust debiasing efficacy when transferred to instruction-fine-tuned models, suggesting that bias representations become entrenched during the pre-training phase, and that both base model and fine-tuning model share these bias representations.

\section{Related Work}
\subsection{Bias and Debiasing in LMs}
The biases of LMs are observed across many independent model outputs and can only be measured by observing the aggregated behavior of LMs \citep{rauh2022characteristics}.  \citet{caliskan2017semantics} proposed the Word Embedding Association Test and discovered that male-associated terms tend to cluster with concepts like work, mathematics, and science, while female-associated terms align more closely with family and artt. \citet{zhao2018gender} introduced a dataset in which only an ambiguous pronoun differs between sentence pairs. LMs must parse the pronoun into one of the two entities mentioned prior to the sentence. Based on this dataset, researchers have demonstrated the widespread existence of gender bias in language models (\citet{touvron2023llama}; \citealp{biderman2023pythia}). \citet{manvi2024large} proposed studying LLMs through the lens of geography and found that LLMs are biased agatinst locations or groups on a variety of sensitive subjective topics. As LLMs' comprehension capabilities continue to enhance, their performance scores on coreference resolution tasks are consistently rising. Consequently, in the future, utilizing such intermediary tasks to evaluate biases may no longer be a viable approach. \citet{nangia2020Crows}, \citet{nadeem2021StereoSet}, \citet{liang2021towards}, \citet{esiobu2023robbie}, and \citet{barikeri2021redditbias} introduced benchmarks to quantify stereotypes across gender, occupation, race, and religion by analyzing language models' probability distributions over stereotype-related target words. To mitigate bias, various debiasing methods are proposed. Re-pre-training with counterfactual data augmentation (\citealp{zmigrod2019counterfactual}) replaces words in the corpus with tuples to construct a counterfactual enhanced corpus for training LMs. \citet{kaneko2021debiasing}, \citet{ravfogel2020null} proposed eliminating biases through projecting bias representations. \citet{schick2021self} and \citet{furniturewala2024thinking} list unexpected behaviors in the prompts and leverage the abilities of the LLM itself to reduce the biased tokens. \citet{banerjee2024all} and \citet{liprompting} generate counterfactuals for different groups, calculate probability distributions using LLMs, and adjust probabilities to produce fairer outputs. \citet{xu2025biasedit} employed model-editing technology to debias LMs.

\subsection{LLM unlearning}
\label{sec:unlearn}
Machine Unlearning \cite{fan2023salun} aims to eliminate the influence of specific training data (such as sensitive or illegal information) on the completed pre-trained model while maintaining the practicability of the model. \citet{yao2024large} applied machine unlearning in the domain of LLMs, referred to as LLM unlearning. LLM unlearning is employed to remove private information, copyrighted content, or harmful data from LLMs, with the goal of erasing specific information while maintaining the model's general performance and functionality  (\citealp{yuan2024closer}; \citealp{fan2023salun}). \citet{liu2025rethinking} suggested that debiasing constitutes one of the primary applications of LLM unlearning.

The most common LLM unlearning method is gradient ascent (GA), which achieves unlearning by maximizing the loss on the forget set \citep{yao2024large}. Some studies introduce Kullback-Leibler (KL) divergence \citep{pan2025multi} and gradient descent (GD) (\citealp{yao2024large}; \citealp{premptis2025ails}) to reduce parameter changes in models before and after unlearning. \citet{zhangnegative} proposed Negative Preference Optimization (NPO), whose progression toward catastrophic collapse is exponentially slower than GA. \citet{chen2023fast} attempted to utilize unlearning for debiasing. However, debiasing usually involves simultaneously mitigating multiple social biases, and the severity of various biases in large models varies \cite{haller2024opiniongpt}. The general unlearning methods cannot simultaneously balance so many biases while still maintaining the ability and knowledge of the model.

\section{Debiasing with BiasUnlearn}

The goal of unlearning knowledge or factual information is to prevent LMs from generating specific content. However, mitigating bias through unlearning does not entail completely severing the association between LMs and specific words. Instead, it focuses on ensuring that the probability of certain words appearing is equalized across different demographic contexts \cite{banerjee2024all}. An ideal debiased language model should achieve an optimal balance between substantial bias mitigation and the preservation of original competencies. Therefore, our proposed BiasUnlearn framework must address three intertwined challenges: (1) ‌Significant bias mitigation; (2) Preserving model capabilities post-debiasing; (3) Avoiding over-debiasing-induced bias reversal.

Consider an LM $\mathcal{M}$ with parameters $\Theta$ and an input context $x$ related to a demographic group $G$, outcome $y=\mathcal{M}(x;\Theta)$, debiasing $\mathcal{M}$ to make the probability of generating a neutral word $w$ equally likely regardless of demographic groups in contexts where only demographic groups differ, such that $\forall$$w \in \mathbb{W}$, ${p(w|x_{i},G_{i},\mathcal{M},\hat{\Theta})=p(w|x_{j},G_{j},\mathcal{M},\hat{\Theta})}$,where $\hat{\Theta}$ is the parameter that has been debiased, that is, $\mathcal{M}$ satisfies equal social group associations.

In this paper, we propose BiasUnlearn (as shown in Figure~\ref{fig:overall}) for mitigating social bias in language models, which incorporates dual-pathway unlearning mechanisms that coordinate stereotype forgetting with anti-stereotype retention.

\vspace{2mm}
\noindent\textbf{Stereotype Forgetting} Firstly, in order to effectively mitigate bias in large models, NPO \citep{zhangnegative} which is more stable than Gradient Ascent is included in the optimization process on stereotypical set ${\mathcal{D}_{s}}$ to forget bias:

\begin{equation}
  \label{eq:eq1}
    \mathcal{L}_{Forget}=-\frac{2}{\beta} \mathbb{E}_{\mathcal{D}_{s}}\left[\log \sigma\left(-\beta \log \frac{\pi_{\theta}(y \mid x)}{\pi_{\mathrm{ref}}(y \mid x)}\right)\right]
\end{equation}

where $\pi_{\theta}$ is the unlearned LM and $\pi_{\mathrm{ref}}$ is the reference LM. $\sigma$ is the sigmoid function and $\beta$ is a regularization parameter.

\vspace{2mm}
\noindent\textbf{Counterfactual Data and Anti-stereotype Retention} For a given context $x$ containing a sensitive word $w_{s}$ related to group $G_{i}$, substituting $w_{s}$ with a non-stereotypical word $w_{a}$ constructs an anti-stereotypical context $x'$, which serves as counterfactual data. 

We adopt various methods to ensure that the language model maintains its general capabilities during unlearning. According to \citet{xu2025zjuklab} and \citet{ji2024reversing}, the integration of forgetting loss with gradient descent optimization objectives contributes to enhanced stability in parameter updates. Inspired by \citet{premptis2025ails}, LMs can remember some facts while forgetting others, we use cross-entropy loss to train LMs with gradient descent optimization on the anti-stereotypical set ${\mathcal{D}_{a}}$, guiding LMs to internalize anti-stereotypical knowledge in parameter space.

\begin{equation}
  \label{eq:example}
   \mathcal{L}_{Retention}=\frac{1}{\left|\mathcal{D}_{a}^{i}\right|} \sum_{\mathcal{D}_{a}^{i}} \mathrm{CE}(y, \hat{y})
\end{equation}

\noindent\textbf{Adversarial Forget Set} Since forgetting loss aims to forget biases, while retention loss focuses on retaining anti-stereotypical information, both objectives align during training, continuous optimization may lead to bias reversal for some bias types. To mitigate this, our experiments demonstrate that incorporating a small portion of anti-stereotypical data into the forgetting training dataset slows down bias reversal and helps prevent language model collapse. Thus the training set $D_{s}$ in formula (\ref{eq:eq1}) is replaced with $D_{s} \cup D_{a'}$, $D_{a'}$ is a subset of $D_{a}$. These anti-stereotypical texts share identical contexts with their stereotypical counterparts except for critical word substitutions, which forces the model to optimize the same set of parameters to accommodate contradictory objectives during training. The gradient interference introduced by counterfactual data also has a regularization effect, preventing parameters from overfitting to a single objective (simply maximizing the loss of stereotypical data).

\vspace{2mm}
\noindent\textbf{Data Chunk} We combine a batch of stereotype data and a batch of anti-stereotype data into a data chunk. According to \citet{premptis2025ails}, after performing the GA step, several consecutive GD annealing steps are required. In BiasUnlearn, we configure the ratio between forgettable and retained data to ${1:n\,(n>1)}$ within each data chunk, and the combined loss function can achieve the same progressive optimization effect as annealing. Since the batch size of the retain set is larger than that of the forget set, during training, retained data is cyclically sampled from the retain set.

\vspace{2mm}
\noindent\textbf{Forward KL Divergence} Following \citet{murphy2022probabilistic} and \citet{yao2024large}, we use forward KL divergence to force the distribution of the unlearned LM to cover all the areas of space of the original LM on the data unrelated to stereotypes, further preventing model collapse:
\begin{equation}
  \label{eq:example}    \mathcal{L}_{KL} = \operatorname{KL}\left( P_{\pi_{\theta}}(x_{unrel}) \parallel P_{\pi_{ref}}(x_{unrel}) \right)
\end{equation}

The final loss is computed as follows:
\begin{equation}
  \label{eq:example}   \mathcal{L}=\alpha_{1}\mathcal{L}_{Forget}+\alpha_{2}\mathcal{L}_{Retention}+\alpha_{3}\mathcal{L}_{KL}
\end{equation}
where alphas are hyperparameters.

\vspace{2mm}
\noindent\textbf{Dynamic Dataset Swapping} To prevent the over-debiasing during unlearning training from leading to bias reversal, we will swap the forget set and the retain set of specific bias types during training based on the results from the development set. Once the SS score for a particular bias type falls below 50, we will exchange the two sets of that bias type to continue training.

In our experiments, we leverage parameter-efficient gradient-based methods \cite{jang2023knowledge} instead of full parameter training to improve training effectiveness. Specifically, we employ low-rank adaptation (LoRA) \cite{hulora}. The specific experimental setups can be found in Appendix \ref{sec:setup}.

\section{Experiments}

\subsection{Dataset and Evaluation Metrics}

\noindent\textbf{StereoSet} \cite{nadeem2021StereoSet} We train the BiasUnlearn model on the StereoSet dataset, which is a large-scale dataset used to measure stereotypes in four domains: gender, profession, race, and religion. Following \citet{xu2025biasedit}, we adopt the test set of StereoSet as our training set and repurpose the development set as our test set. The trained model will also be evaluated on other benchmarks mentioned later. More details about the training set can be found in Appendix \ref{sec:dataset}.

StereoSet evaluates LMs through three metrics: StereoSet (SS) score, Language Modeling (LM) score, and Idealized Context Correlation Test (ICAT) score. SS score represents the percentage of instances in which an LM prefers stereotypical sentences to anti-stereotypical sentences. The SS score approximating 50 indicates optimal bias fairness. The LM score is used to eval language modeling and general capabilities of an LM, that is, LM prioritizes the percentage of instances with meaningful associations rather than irrelevant alternatives. The ICAT score integrates SS score and LM score, higher values is better, calculates as follows:
\begin{equation}
  \label{eq:example}
   ICAT=\operatorname{LMS} \frac{\min (SS, 100-SS)}{50}
\end{equation}

\noindent\textbf{Crows-Pairs} \cite{nangia2020Crows} is a crowdsourced benchmark for stereotypical bias analysis. This dataset contains 1,508 examples including nine categories of social biases, such as gender, race, etc. Crows-Pairs is constructed in the form of contrastive pairs: each instance contains two semantically similar sentences, one of which presents a stereotypical description of historically disadvantaged groups, and the other is a contrasting expression of anti-stereotypes. Following \citet{banerjee2024all}, we also adopt the StereoSet score to  measure the gender, religion, and race bias in LMs, whose ideal score is 50\%.


\begin{table*}[htbp]
\centering
\renewcommand{\arraystretch}{1.2}
\scalebox{0.72}{
\begin{tabular}{lcccccccccccccc}
\hline
\multicolumn{1}{c}{\multirow{3}{*}{Method}} & \multicolumn{7}{c}{\textbf{GPT2-Medium}}                                                                                                                                                                                                                              & \multicolumn{7}{c}{\textbf{GPT2-Large}}                                                                                                                                                                                                                               \\ \cline{2-15} 
\multicolumn{1}{c}{}                        & \multicolumn{5}{c}{\textbf{Stereotype Score (\%) $\rightarrow$ 50}}                                                                                               & \multirow{2}{1cm}{\centering \textbf{$\Delta$LMS} \\ $\rightarrow$ 0} & \multirow{2}{*}{\textbf{ICAT $\uparrow$}} & \multicolumn{5}{c}{\textbf{Stereotype Score (\%) $\rightarrow$ 50}}                                                                                               & \multirow{2}{1cm}{\centering \textbf{$\Delta$LMS} \\ $\rightarrow$ 0} & \multirow{2}{*}{\textbf{ICAT $\uparrow$}} \\ \cline{2-6} \cline{9-13}
\multicolumn{1}{c}{}                        & \multicolumn{1}{l}{Gend.} & \multicolumn{1}{l}{Prof.} & \multicolumn{1}{l}{Race} & \multicolumn{1}{l}{Reli.} & \multicolumn{1}{l}{\textcolor{red}{Overall}} &                                          &                                           & \multicolumn{1}{l}{Gend.} & \multicolumn{1}{l}{Prof.} & \multicolumn{1}{l}{Race} & \multicolumn{1}{l}{Reli.} & \multicolumn{1}{l}{\textcolor{red}{Overall}} &                                          &                                           \\
BaseModel                                   & 65.58                     & 63.37                     & 61.44                    & 62.57                     & 62.74                                                           & 92.21                                    & 68.71                                     & 65.29                     & 65.68                     & 63.0                     & 61.61                     & 64.26                                                           & 92.49                                    & 66.12                                     \\ \hline
CDA                                         & 63.72                     & 58.42                     & 48.41                    & 51.68                     & 54.27                                                           & 2.98                                     & 87.05                                     & 67.07                     & 54.25                     & 47.24                    & \textbf{51.68}            & 52.58                                                           & 3.55                                     & \textbf{91.09}                            \\
Self-Debias                                 & 59.63                     & 61.35                     & 57.5                     & 57.98                     & 59.25                                                           & -3.27                                    & 72.48                                     & 65.09                     & 64.34                     & 57.51                    & 57.98                     & 61.08                                                           & -3.27                                    & 69.44                                     \\
CAFIE                                       & 56.2                      & 62.03                     & 60.66                    & 63.91                     & 60.74                                                           & -3.14                                    & 69.94                                     & 59.15                     & 62.74                     & 60.84                    & 61.61                     & 61.38                                                           & -3.02                                    & 69.11                                     \\
BiasEdit                                    & \textbf{49.42}            & 56.25                     & 52.38                    & 54.33                     & 53.87                                                           & -2.63                                    & 82.66                                     & \textbf{52.64}            & 55.27                     & 54.02                    & 47.36                     & 54.19                                                           & -3.13                                    & 82.8                                      \\
\textbf{BiasUnlearn}                        & 52.44                     & \textbf{51.06}            & \textbf{50.37}           & \textbf{48.64}            & \textbf{50.83}                                                  & \textbf{-0.2}                            & \textbf{90.48}                            & 53.9                      & \textbf{52.79}            & \textbf{48.17}           & 47.49                     & \textbf{50.62}                                                  & \textbf{-0.26}                           & 91.08                                     \\ \hline
\multicolumn{1}{c}{\multirow{3}{*}{Method}} & \multicolumn{7}{c}{\textbf{Mistral-7B}}                                                                                                                                                                                                                               & \multicolumn{7}{c}{\textbf{Llama3-8B}}                                                                                                                                                                                                                                \\ \cline{2-15} 
\multicolumn{1}{c}{}                        & \multicolumn{5}{c}{\textbf{Stereotype Score (\%) $\rightarrow$ 50}}                                                                                               & \multirow{2}{1cm}{\centering \textbf{$\Delta$LMS} \\ $\rightarrow$ 0} & \multirow{2}{*}{\textbf{ICAT $\uparrow$}} & \multicolumn{5}{c}{\textbf{Stereotype Score (\%) $\rightarrow$ 50}}                                                                                               & \multirow{2}{1cm}{\centering \textbf{$\Delta$LMS} \\ $\rightarrow$ 0} & \multirow{2}{*}{\textbf{ICAT $\uparrow$}} \\ \cline{2-6} \cline{9-13}
\multicolumn{1}{c}{}                        & \multicolumn{1}{l}{Gend.} & \multicolumn{1}{l}{Prof.} & \multicolumn{1}{l}{Race} & \multicolumn{1}{l}{Reli.} & \multicolumn{1}{l}{\textcolor{red}{Overall}} &                                          &                                           & \multicolumn{1}{l}{Gend.} & \multicolumn{1}{l}{Prof.} & \multicolumn{1}{l}{Race} & \multicolumn{1}{l}{Reli.} & \multicolumn{1}{l}{\textcolor{red}{Overall}} &                                          &                                           \\
BaseModel                                   & 69.83                     & 63.09                     & 66.31                    & 57.24                     & 65.19                                                           & 92.26                                    & 64.23                                     & 75.29                     & 68.4                      & 65.24                    & 64.69                     & 67.69                                                           & 94.08                                    & 60.8                                      \\ \hline
CDA                                         & 51.98                     & 47.01                     & 47.22                    & 45.41                     & 47.56                                                           & 1.69                                     & 91.67                                     & 69.05                     & 50.44                     & 54.07                    & 55.66                     & 55.18                                                           & 4.32                                     & 85.85                                     \\
Self-Debias                                 & 62.15                     & 55.77                     & 50.86                    & 59.72                     & 54.49                                                           & -32.97                                   & 51.37                                     & 64.7                      & 57.56                     & 55.78                    & 52.28                     & 57.45                                                           & -35.82                                   & 52                                        \\
CAFIE                                       & 57.61                     & 62.85                     & 67.55                    & \textbf{51.13}            & 63.89                                                           & -5.43                                    & 62.95                                     & 55.41                     & 65.7                      & 66.01                    & 61.24                     & 64.37                                                           & -5.11                                    & 63.17                                     \\
BiasEdit                                    & 46.66                     & 45.21                     & 46.06                    & 51.14                     & 45.93                                                           & -13.56                                   & 60.81                                     & 44.73                     & 40.32                     & \textbf{50.49}           & 52.01                     & 45.2                                                            & -26.07                                   & 72.79                                     \\
\textbf{BiasUnlearn}                        & \textbf{51.4}             & \textbf{48.36}            & \textbf{50.67}           & 51.49                     & \textbf{49.92}                                                  & \textbf{-1.42}                           & \textbf{91.71}                            & \textbf{54.69}            & \textbf{49.99}            & 52.47                    & \textbf{49.98}            & \textbf{51.71}                                                  & \textbf{-0.4}                            & \textbf{89.09}                            \\ \hline
\end{tabular}}
  \caption{Debiasing performance of BiasUnlearn compared to baselines on StereoSet. Overall represents the aggregated results across all bias types. The SS score should approach 50 optimally, while the LM score requires minimal deviation from the base model. The ICAT metric theoretically exhibits positive correlation with performance improvement.}
\label{tab:t1}
\end{table*}

\begin{table}[h]
\centering
\renewcommand{\arraystretch}{1.2}
\scalebox{0.72}{
\begin{tabular}{lcccccc}
\hline
\multicolumn{1}{c}{\multirow{2}{*}{Method}} & \multicolumn{3}{c}{\textbf{GPT2-Medium $\rightarrow$ 50}}                                         & \multicolumn{3}{c}{\textbf{GPT2-Large $\rightarrow$ 50}}                                          \\ \cline{2-7} 
\multicolumn{1}{c}{}                        & \multicolumn{1}{l}{Gend.} & \multicolumn{1}{l}{Race} & \multicolumn{1}{l}{Reli.} & \multicolumn{1}{l}{Gend.} & \multicolumn{1}{l}{Race} & \multicolumn{1}{l}{Reli.} \\
BaseModel                                   & 59.16                     & 62.4                     & 72.38                     & 59.16                     & 62.21                    & 71.43                     \\
CDA                                         & 59.54                     & 50.31                    & 65.71                     & 59.94                     & 58.49                    & 71.43                     \\
Self-Debias                                 & \textbf{49.62}            & 46.54                    & 51.43                     & 55.73                     & 58.49                    & 60                        \\
CAFIE                                       & 46.18                     & 47.17                    & \textbf{50.48}            & \textbf{50}               & 54.26                    & 59.05                     \\
BiasEdit                                    & 56.08                     & 52.66                    & 54.66                     & 51.36                     & 47.5                     & 46.92                     \\
\textbf{BiasUnlearn}                        & 52.31                     & \textbf{50}            & 47.97                     & 52.96                     & \textbf{50.04}           & \textbf{49.52}            \\ \hline
\multicolumn{1}{c}{\multirow{2}{*}{Method}} & \multicolumn{3}{c}{\textbf{Mistral-7B $\rightarrow$ 50}}                                          & \multicolumn{3}{c}{\textbf{Llama3-8B $\rightarrow$ 50}}                                           \\ \cline{2-7} 
\multicolumn{1}{c}{}                        & \multicolumn{1}{l}{Gend.} & \multicolumn{1}{l}{Race} & \multicolumn{1}{l}{Reli.} & \multicolumn{1}{l}{Gend.} & \multicolumn{1}{l}{Race} & \multicolumn{1}{l}{Reli.} \\
BaseModel                                   & 62.98                     & 54.72                    & 69.52                     & 60.31                     & 59.12                    & 74.29                     \\
CDA                                         & 58.78                     & 54.09                    & 71.43                     & 40.08                     & 55.35                    & \textbf{51.43}            \\
Self-Debias                                 & 54.2                      & 55.35                    & 55.24                     & 58.78                     & 61.64                    & 64.76                     \\
CAFIE                                       & 42.37                     & 48.43                    & 53.33                     & 42.75                     & 49.06                    & 60.95                     \\
BiasEdit                                    & 51.46                     & 41.49                    & 46.73                     & 46.04                     & \textbf{50}              & 55.9                      \\
\textbf{BiasUnlearn}                        & \textbf{51.15}            & \textbf{47.17}           & \textbf{51.43}            & \textbf{49.62}            & 50.31                    & 52.38                     \\ \hline
\end{tabular}}
  \caption{Debiasing performance of BiasUnlearn compared to baselines on Crows-Pairs. The SS score (\%) should approach 50 optimally.}
\label{tab:t2}
\end{table}

\noindent\textbf{BBQ, GLUE, FLUTE, AmazonPolarity, MT-bench} For fine-tuned LLMs, we employ BBQ \cite{parrish-etal-2022-bbq}, four tasks of GLUE\cite{wangglue}, FLUTE \cite{chakrabarty-etal-2022-flute}, AmazonPolarity \cite{enevoldsen2025mmtebmassivemultilingualtext}, MT-bench \cite{zheng2023judging}, CEB \cite{wangceb} and FairMT \cite{fanfairmt} for evaluation. Details are in Appendix \ref{sec:dataset}.


\subsection{Baselines}

We compare BiasUnlearn with the following four baselines: counterfactual
 factual data augmentation (CDA) \cite{zmigrod2019counterfactual}, Self-Debias \cite{schick2021self}, Counterfactually Aware Fair Inference (CAFIE) \cite{banerjee2024all}, and BiasEdit \cite{xu2025biasedit}. Since we need to deal with multiple types of biases simultaneously, we use the anti-stereotypical data from StereoSet as counterfactual data for training in our implementation CDA. For Self-Debias, we use the implementation of \cite{meade2022empirical}. As for CAFIE and BiasEdit, we use their official implementations and hyperparameters. More details are in Appendix \ref{sec:baselines}.

As a model-agnostic debiasing method, BiasUnlearn can be applied to any language model. We conduct experiments on GPT2-medium (355M), GPT2-Large (774M) \cite{radfordlanguage}, Mistral-7B \cite{jiang2023mistral7b}, and Llama3-8B \cite{meta2024introducing}, which represent diverse language models across different sizes. To verify the generalization of our method, we also conducted further experiments with the instruction-fine-tuned models of Mistral-7B and Llama3-8B.

\subsection{Results}

\begin{table*}[htbp]
\renewcommand{\arraystretch}{1.2}
\scalebox{0.71}{
\begin{tabular}{lcccccccccccccc}
\hline
\multicolumn{1}{c}{\multirow{3}{*}{Method}} & \multicolumn{7}{c}{\textbf{Mistral-7B-Instruction}}                                                                                                                                                                               & \multicolumn{7}{c}{\textbf{Llama3-8B-Instruction}}                                                                                                                                                                                \\ \cline{2-15} 
\multicolumn{1}{c}{}                        & \multicolumn{5}{c}{\textbf{\textbf{Stereotype Score (\%) $\rightarrow$ 50}}}                                                           & \multirow{2}{*}{\textbf{LMS $\uparrow$}} & \multirow{2}{*}{\textbf{ICAT $\uparrow$}} & \multicolumn{5}{c}{\textbf{\textbf{Stereotype Score (\%) $\rightarrow$ 50}}}                                                           & \multirow{2}{*}{\textbf{LMS $\uparrow$}} & \multirow{2}{*}{\textbf{ICAT $\uparrow$}} \\ \cline{2-6} \cline{9-13}
\multicolumn{1}{c}{}                        & \multicolumn{1}{l}{Gend.} & \multicolumn{1}{l}{Prof.} & \multicolumn{1}{l}{Race} & \multicolumn{1}{l}{Reli.} & \multicolumn{1}{l}{Overall} &                                          &                                           & \multicolumn{1}{l}{Gend.} & \multicolumn{1}{l}{Prof.} & \multicolumn{1}{l}{Race} & \multicolumn{1}{l}{Reli.} & \multicolumn{1}{l}{Overall} &                                          &                                           \\
BaseModel                                   & 72.16                     & 63.70                     & 65.54                    & 53.79                     & 65.23                       & \textbf{89.92}                           & 62.52                                     & 68.15                     & 66.54                     & 65.44                    & 61.24                     & 66.04                       & \textbf{91.06}                           & 61.85                                     \\
BiasUnlearn                                 & \textbf{53.07}            & 46.92                     & \textbf{50.05}           & \textbf{48.46}            & \textbf{49.18}              & 89.03                                    & \textbf{87.57}                            & \textbf{49.14}            & \textbf{48.87}            & \textbf{47.95}           & \textbf{47.31}            & 48.42                       & 91.03                                    & 88.16                                     \\
WeightTrans                                 & 57.27                     & \textbf{52.35}            & 54.13                    & 47.68                     & 53.61                       & 89.90                                    & 83.41                                     & 51.64                     & 47.62                     & 53.51                    & 45.98                     & \textbf{50.75}              & 89.82                                    & \textbf{88.47}                            \\ \hline
\end{tabular}}
  \caption{Debiasing performance comparison of instruction-fine-tuned models: Before vs. after applying BiasUnlearn for training or debiasing weight transfer.}
\label{tab:t3}

\end{table*}

\begin{table*}[htbp]
\renewcommand{\arraystretch}{1.2}
\scalebox{0.73}{
\begin{tabular}{lcccccccccccccc}
\hline
\multicolumn{1}{c}{\multirow{2}{*}{\textbf{Method}}} & \multicolumn{7}{c}{\textbf{Mistral-7B-Instruction}}                                                                                   & \multicolumn{7}{c}{\textbf{Llama3-8B-Instruction}}                                                               \\ \cline{2-15} 
\multicolumn{1}{c}{}                                 & \textbf{SST}  & \textbf{MRPC} & \textbf{COLA} & \textbf{RTE}  & \textbf{Amazon} & \textbf{Flute} & \multicolumn{1}{c|}{\textbf{MT}}   & \textbf{SST}  & \textbf{MRPC} & \textbf{COLA} & \textbf{RTE}  & \textbf{Amazon} & \textbf{Flute} & \textbf{MT}   \\
BaseModel                                            & 0.94          & \textbf{0.73} & 0.74          & 0.23          & \textbf{0.91}   & 0.85           & \multicolumn{1}{c|}{6.60}          & 0.93          & 0.68          & 0.74          & 0.22          & 0.89            & \textbf{0.41}  & \textbf{7.83} \\
BiasUnlearn                                          & 0.94          & 0.71          & 0.74          & 0.24          & 0.90            & 0.86           & \multicolumn{1}{c|}{6.49}          & \textbf{0.94} & 0.67          & 0.72          & \textbf{0.27} & \textbf{0.90}   & 0.38           & 7.68          \\
WeightTrans                                          & \textbf{0.94} & 0.72          & \textbf{0.75} & \textbf{0.24} & 0.90            & \textbf{0.86}  & \multicolumn{1}{c|}{\textbf{6.76}} & 0.93          & \textbf{0.68} & \textbf{0.74} & 0.25          & 0.89            & 0.40           & 7.71          \\ \hline
\end{tabular}}
  \caption{Performance comparison of instruction-fine-tuned models in semantic understanding tasks as well as question-answer dialogue task: Before vs. after applying BiasUnlearn for training or debiasing weight transfer. In the GLUE, FLUTE, and AmazonPolarity benchmarks, F1 scores are reported. We utilize Llama-3-70B-Instruction to evaluate the responses of the assessed LMs in MT-bench by assigning scores.}
\label{tab:t4}
\end{table*}

\noindent\textbf{BiasUnlearn demonstrates superior debiasing performance while maintaining language modeling scores}. Firstly, BiasUnlearn achieves significant reductions in stereotyping scores without excessive debiasing, according to Table~\ref{tab:t1}, the overall SS scores of BiasUnlearn decreased by at least 11.91 and at most 15.98 on the four models. And BiasUnlearn can reduce the SS scores of all bias types to around 50\%, while the absolute values of the difference between SS value and 50\% of baselines are mostly higher than BiasUnlean. From the results of the baseline methods, we observe that most approaches exhibit debiasing effects, however, the LMS values also significantly decreased. Conversely, since unlearning training will not improve the model's general ability or increase its knowledge, an increase in the LMS indicates that the model overfits to specific data, which is also detrimental to the performance of LMs. Compared with other methods, BiasUnlean achieved the minimal change on LMS (The minimum change of BiasUnlearn is only 0.2 and the maximum is only 1.42), indicating that BiasUnlearn method does not adversely affect the language modeling capabilities of the original models and our method maximally preserves the capabilities of the original models.

Based on Table~\ref{tab:t1} and Table~\ref{tab:t2}, which present the results of various debiasing methods on StereoSet and Crows-Pairs, substantial differences in the distribution between the two datasets are evident. In fact, StereoSet divides race by nationality while Crows-Pairs divides race based on skin color. BiasUnlearn achieves significant reductions in SS scores on both StereoSet and Crows-Pairs with the same training set, demonstrating its excellent debiasing effects and robust generalization capability. Prompt-based methods like Self-Debias and CAFIE struggle to adapt to different data distributions due to vocabulary limitations. Theoretically, fine-tune-based methods exhibit better adaptability across different datasets, but they relies heavily on appropriate training data. Notably, since the data distribution varies even across categories within the StereoSet dataset itself, fine-tuning approaches like CDA demonstrate strong performance on specific datasets or categories but suffer from poor generalization to others. BiasEdit achieved significant debiasing effect on various datasets and bias types, however, after editing, the LMS of all models has significantly decreased, which negatively impacts the overall performance of the language models.

\begin{figure*}[htbp]
  \includegraphics[width=0.475\linewidth]{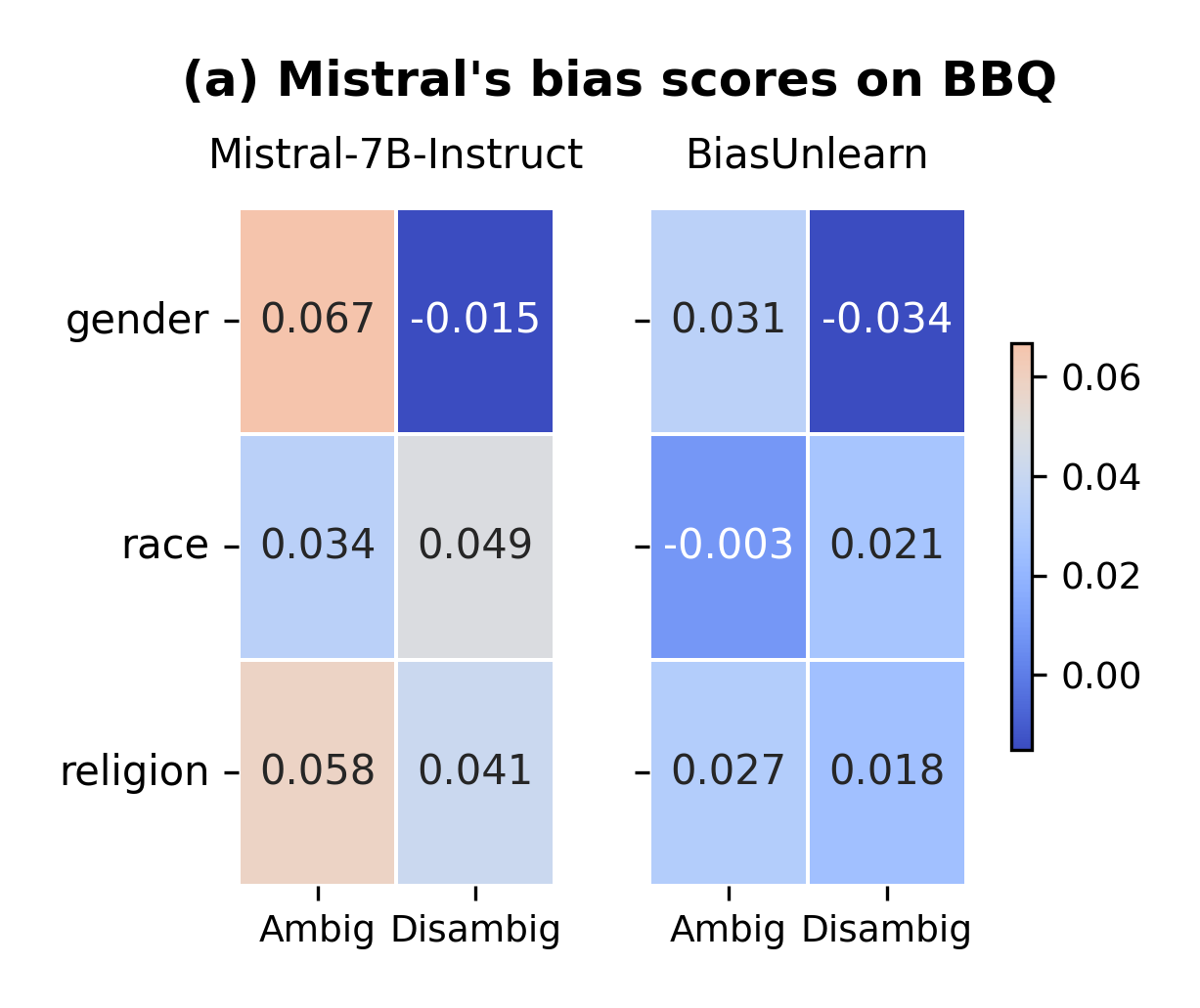} \hfill
  \includegraphics[width=0.485\linewidth]{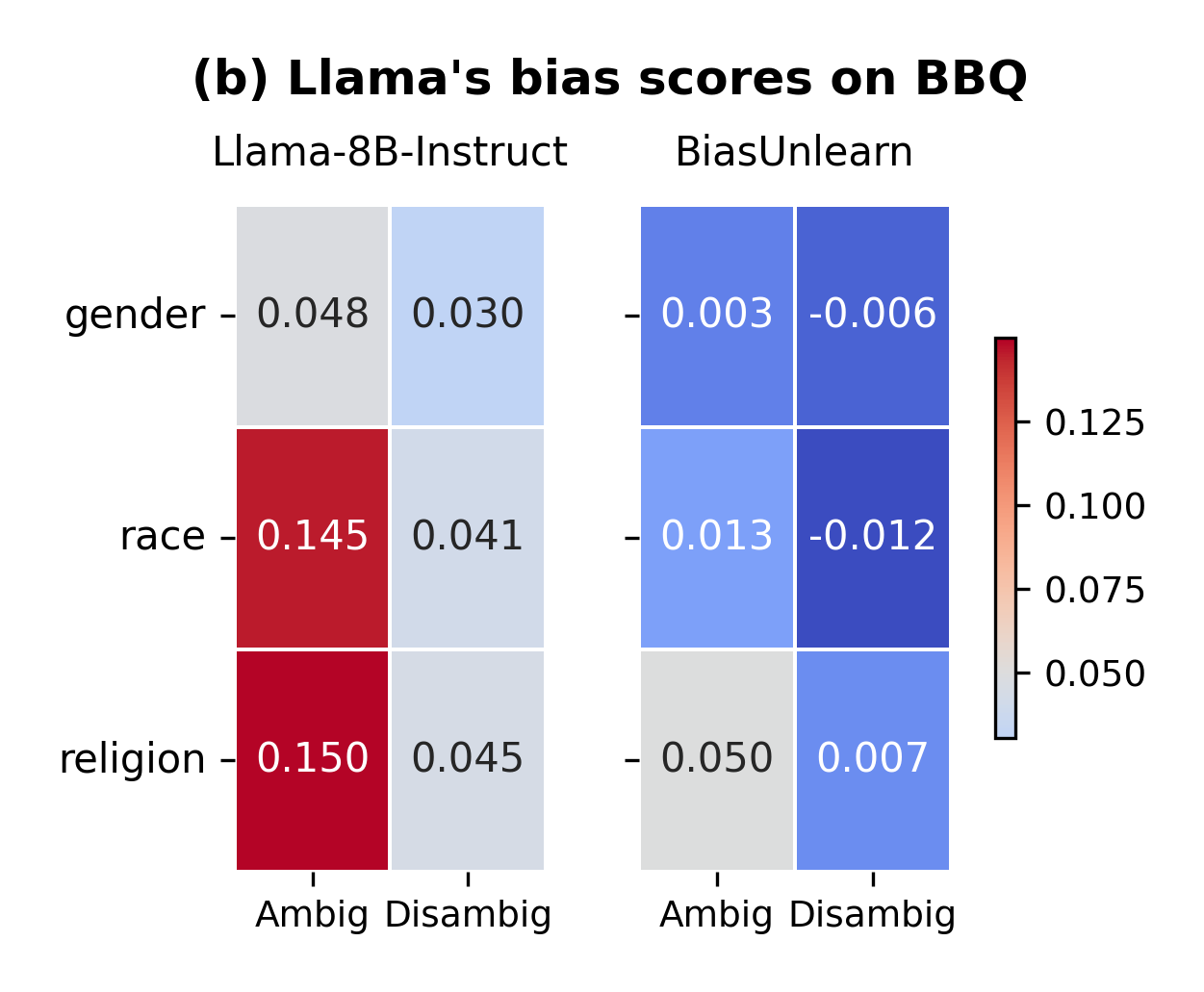}
  \caption {Bias scores in each category of BBQ, split by whether the context is ambiguous or disambiguated. The higher the bias score, the stronger the bias.}
  \label{fig:bbq}
\end{figure*}

\vspace{2mm}
\noindent\textbf{BiasUnlearn works for both pretrained and fine-tuned models} We conducted further experiments on downstream tasks on the fine-tuned models, and the results are presented in Table~\ref{tab:t4} and Figure~\ref{fig:bbq}. We can see that after training with BiasUnlearn, the model achieved performance levels that closely mirrored those of the original models across various tasks. This finding provides further evidence that our strategy for debiasing while retaining model capabilities is highly effective.

From Figure~\ref{fig:bbq}, we can observe that after BiasUnlearn training, the instruction-fine-tuned model demonstrates significant bias reduction in QA tasks. However, the bias scores in ambiguous contexts are much higher, indicating that when there is no clear answer, LMs will rely on social stereotypes. Notably, LLMs exhibit lower bias scores on the BBQ benchmark, which can be attributed to their enhanced general capabilities that enable answers to be less dependent on stereotypical associations. The improved capacity for context-aware reasoning consequently increases answer accuracy while reducing bias metrics.
This finding indirectly suggests that employing intermediate evaluation tasks like QA formats and questionnaires for bias evaluation requires more nuanced design considerations. Current implementations may lack the sophistication needed to effectively probe deeper, more systemic biases within large language models (\citealp{shrawgi2024uncovering}; \citealp{duandenevil}).

\vspace{2mm}
\noindent\textbf{The transferability of debiasing weights indicates that bias features have been solidified in the pre-training stage.} To further verify that the parameters learned by BiasUnlearn are only related to stereotypes, we load the debiasing weights learned on base models into instruction-finetuned models. As shown in Table~\ref{tab:t3}, both instruction-fine-tuned models of Llama and Mistral achieve strong debiasing effects after loading the base model's debiasing weights, with only minor degradation in language modeling scores. And from Table~\ref{tab:t4}, the instruction-finetuned models with transferred debiasing weights exhibit almost no performance drop on GLUE and other tasks compared to the original models. It indicates that bias features are deeply solidified within the underlying features during the pre-training phase and cannot be modified by ordinary fine-tuning. Furthermore, the successful transfer of debiasing weights indicates that BiasUnlearn effectively disentangles bias-related parameters without reconstructing the underlying semantic representations. 

The previous works (\citealp{kadhe2024split}; \citealp{zhang2023composing}) divided the training data into specific behavioral subtypes that were separately trained through separate LLMs based on attribute values, and then merged them into the final model. Our experiments are significantly different from their work, as they combine multiple weights from the same LM, while we transfer the same weights from different models. Our findings suggest that in the future applications of large language models, there will be no need to debias each individual model: once an unbiased base model is obtained, its subsequent fine-tuning models will also be essentially unbiased; and by only debiasing biased base models and transferring the debiasing weights to their fine-tuned large models that need to be applied, the workload can be significantly reduced.

\noindent\textbf{BiasUnlearn is  efficient, and the resulting debiased models can be deployed effortlessly}. During inference, as the number of BiasUnlearn parameters does not change significantly, the inference speed will not slow down and memory consumption will not significantly increase; while prompt-based methods require parallel or sequential multi-step reasoning, which will increase memory usage or inference time by several times. Compared to fine-tuning approaches such as BiasEdit, which require multiple epochs to train, BiasUnlearn can complete training within merely a few hundred steps, significantly reducing the training time required. Additionally, BiasUnlearn does not require any modifications to the inference process (such as adding sensitive word lists, or modifying token distribution during inference, etc.), so that the debiased LMs can be deployed like vanilla LMs. Compared to BiasEdit, which is only evaluated on sentence-level examples \cite{xu2025biasedit}, LMs trained by BiasUnlearn are capable of adapting to a variety of tasks (details are in Appendix \ref{sec:dataset}).


\begin{table}[]
\centering
\begin{tabular}{lccc}
\hline
Method             & \textbf{SS}    & \textbf{LMS}                       & \textbf{ICAT}                      \\ \hline
BiasUnlearn        & 53.73          & \multicolumn{1}{l}{\textbf{92.95}} & \multicolumn{1}{l}{\textbf{86.01}} \\
w/o Retention loss & \textbf{47.56} & 57.72                              & 54.91                              \\
w/o KL             & 43.14          & 88.22                              & 76.11                              \\
w/o Adv            & 44.72          & 86.84                              & 77.67                              \\ \hline
\end{tabular}
  \caption{The overall SS, LMS, ICAT of BiasUnlearn with or without $\mathcal{L}_{Retention}$, $\mathcal{L}_{KL}$ or Adversarial Forget Set.}
\label{tab:t5}
\end{table}

\begin{figure}[ht]
  \includegraphics[width=0.98\linewidth]{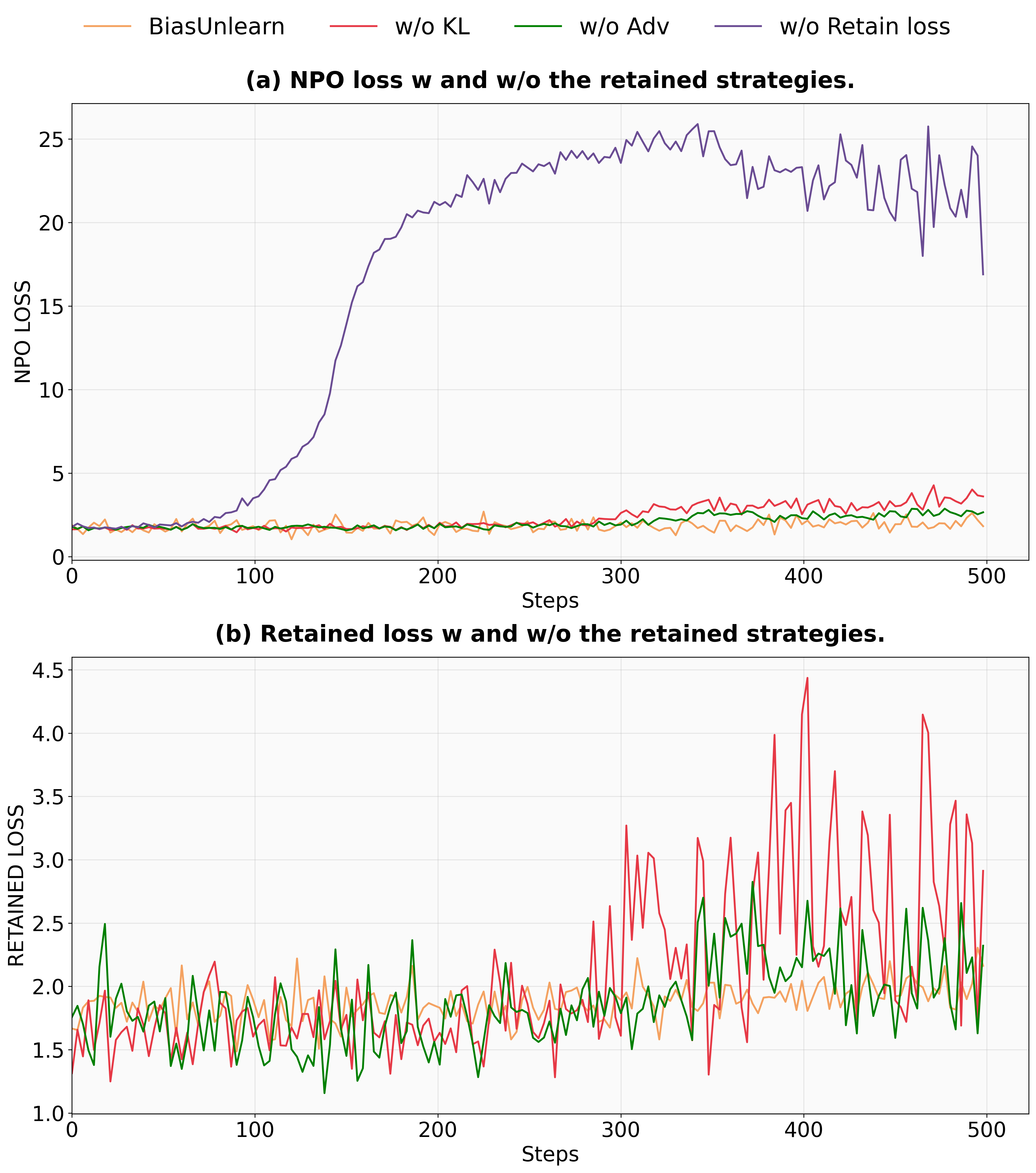} \hfill
  \caption {Forgetting loss and Retention loss of BiasUnlearn with or without $\mathcal{L}_{Retention}$, $\mathcal{L}_{KL}$ or Adversarial Forget Set.}
    \label{fig:ab}
\end{figure}

\subsection{Ablation Experiments}
In order to investigate the influence of different components within the BiasUnlearn framework on retaining the inherent capabilities of language models, we conduct ablation experiments on StereoSet based on GPT2-large. For a fair comparison, we use the checkpoint at step 1000 to generate the evaluation results. As shown in Table~\ref{tab:t5}, even after a thousand steps of unlearning training, the LMS still maintains a high value. Without Retention loss, LMS decreased by 35.23 points, which is the largest decrease among all methods, indicating that retention loss plays the greatest role in maintaining the original abilities of the model during the unlearn process. And under the absence of KL divergence and Adversarial Forget
Set, the LMS exhibits respective reductions of 4.73 and 6.1 points, empirically validating that both KL divergence constraints and Adversarial Forget Set are critical for mitigating catastrophic ability degradation and knowledge forgetting during LLM unlearning. When Adversarial Forget
Setis absent, the SS score decreases lower than when lacking KL divergence, so Adversarial Forget Set has a certain effect on preventing over-debiasing.

From Figure~\ref{fig:ab} we can clearly observe the differential effects of each constituent on unlearning training. Whether the Forgetting loss or Retention loss becomes too large, it is a disaster for bias unlearning, which compromises the stability of the training. When there is a lack of Retention loss, the Forgetting loss starts to grow rapidly around the 100th step, with an absolute value that increases more than twenty times compared to the beginning, causing the collapse of the debiased model. The negative impacts of KL divergence absence on the language model manifest much later than those from Retention loss deterioration. Both Forgetting loss and Retention loss exhibit a smaller increase compared to scenarios lacking Retention loss. The two loss curves without KL divergence are both above the curve with missing Adversarial Forget Set, indicating that KL divergence has a greater impact on the results than Adversarial Forget Set. 

\subsection{Compared with SFT and DPO}

\begin{figure}[ht]
  \includegraphics[width=0.98\linewidth]{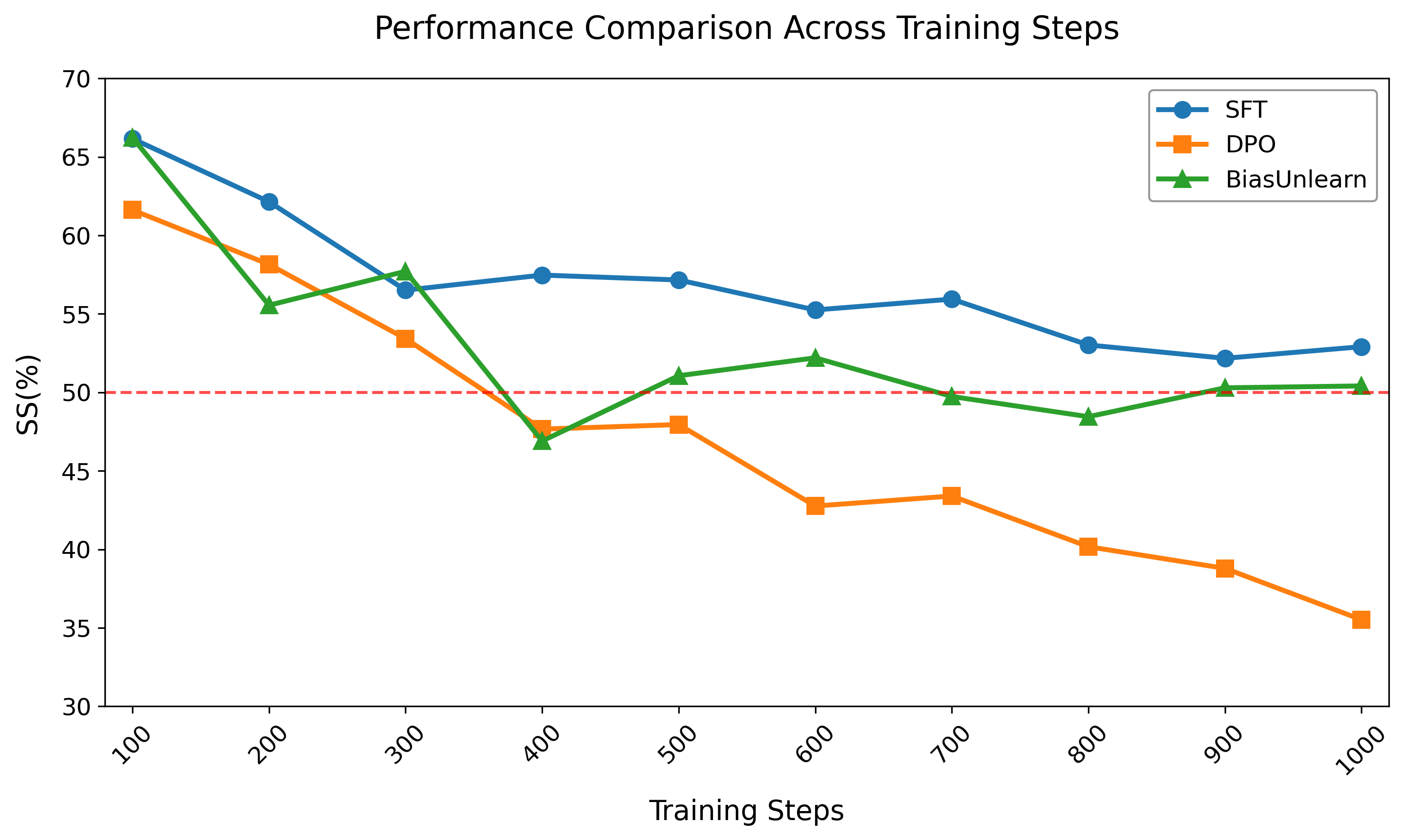} \hfill
  \caption {Comparison results of BiasUnlearn with SFT and DPO.}
    \label{fig:sft}
\end{figure}

We conducted SFT and DPO training on Llama3-8B using the same training set of BiasUnlearn.We only report the overall SS metric since SFT or DPO training on a single type of corpus may compromise the generalization capability of LLMs. The comparison results are shown in Figure~\ref{fig:sft}.

It can be seen that with the increase in training steps, it is difficult for the SS value of SFT to drop below 50. This is because for fairness, the ratio of stereotype to anti-stereotype data in SFT training data is 1:1, so the debiasing effect is not as good as DPO and BiasUnlearn. The SS value of DPO decreases continuously with the increase of steps, therefore, using DPO allows the SS value approach 50, but requires precise hyperparameters and control, otherwise bias will reverse. In fact, the training data contains various biased data, and the severity of various biases in large models varies. Therefore, it is difficult to determine the proportion of various training data, and it is difficult to intervene in DPO training. BiasUnlearn has implemented effective strategies to prevent bias reversal, ensuring that the SS values for various types of biases can be maintained around 50 without the need for hyperparameter tuning.

\subsection{Case Study}

We compared the results generated by the models before and after debiasing on the BBQ and MT-bench, as detailed in Appendix \ref{sec:mtcase}.

\section{Conclusion}

In this work, we propose BiasUnlearn, an effective unlearning approach for debiasing language models. BiasUnlearn incorporates dual-pathway unlearning mechanisms that coordinate stereotype forgetting with anti-stereotype retention. Through adversarial forget set and dynamic dataset swapping, our approach safeguards against bias polarity reversal.  Experimental results demonstrate that BiasUnlearn successfully mitigates bias in language models without reversing the bias direction, while simultaneously maintaining the model's general capabilities. Transfer experiments reveal that our work is transferable across model variants, empirically confirming that bias representations become entrenched during pre-training and persist through fine-tuning phases. These findings not only offer practical debiasing tools but also provide theoretical insights into bias propagation in language models; and this work makes a significant contribution to promoting fairness in LLMs. In the future, we will extend BiasUnlearn to other forms of bias beyond social biases and investigate the relationship between bias representations and different pre-training objectives.

\section*{Limitations}

Existing datasets related to social bias fail to cover a comprehensive range of demographic groups, and many countries, ethnicities, and religions are entirely absent in these datasets. Consequently, language models trained on these datasets using BiasUnlearn may not effectively eliminate bias for all demographic groups. In addition, many types of bias have potential correlations. For example, race and religion exhibit statistical correlations in real-world issues (e.g., highly overlapping racial and religious groups in certain regions) \cite{perry2019christian}, and models may simultaneously affect both types of bias by adjusting shared underlying features. Our work does not fully disentangle the representation space of different biases. When optimizing for a certain type of bias, the parameters associated with other related biases are adjusted jointly.

\section*{Ethics Statement}

Our work effectively removes social biases in language models while maintaining their general capabilities, greatly improving the fairness of language models for different demographic groups. We strictly adhere to the data usage policies of various open-source datasets. The opinions and findings contained in the dataset samples we provide should not be interpreted as representing the views expressed or implied by the authors. 

\section*{Acknowledgments}
This research is supported by Artificial Intelligence National Science and Technology Major Project 2023ZD0121200 and the National Science Fund for Excellent Young Scholars under Grant 62222212.

\bibliography{custom}

\appendix

\section{Experimental Details}
\label{sec:exp}

\subsection{Dataset and Evaluation Metrics}
\label{sec:dataset}

\noindent\textbf{Stereoset} \citealp{nadeem2021StereoSet} Each sample in Stereoset consists of one stereotype sentence, one anti-stereotype sentence, and one sentence unrelated to stereotypes. We adopt the test set of StereoSet as our training set and repurpose the development set as our test set. The number of samples corresponding to each bias type in the re-segmented dataset is presented in  Table~\ref{tab:stere}.

\begin{table}[htbp]
\centering
\renewcommand{\arraystretch}{1.2}
\scalebox{0.8}{
\begin{tabular}{lccccc}
\hline
\textbf{} & \textbf{gender} & \textbf{profession} & \textbf{race} & \textbf{religion} & Overall \\ \hline
train     & 1471            & 4782                & 5871          & 438               & 12562            \\
dev       & 50              & 50                  & 50            & 50                & 200              \\
test      & 497             & 1638                & 1938          & 159               & 4232             \\ \hline
\end{tabular}}
  \caption{The number of samples corresponding to each bias type in our dataset.}
\label{tab:stere}
\end{table}

\vspace{2mm}
\noindent\textbf{BBQ} \cite{parrish-etal-2022-bbq} is a question-answering dataset designed to assess bias across nine social groups. Each question in BBQ is presented in two forms: one is an ambiguous version, lacking clear context, as well as a disambiguation version that provides additional context before the question. BBQ defines the Bias Score as a metric to quantify the degree of bias present in the responses of large language models. A higher score indicates a more severe level of bias.

\vspace{2mm}
 
\noindent\textbf{GLUE} \cite{wangglue} GLUE benchmark comprises nine NLU tasks for evaluating the semantic understanding ability of language models, and We selected a subset of four tasks—SST-2, MRPC, COLA, and RTE to assess model performance. \textbf{SST-2} is an sentiment analysis (positive/negative) task. \textbf{MRP} is a sentence semantic matching task, which involves determining whether two sentences are semantically similar. \textbf{COLA} is a grammar judgment task, which involves determining whether a sentence is grammatically correct. \textbf{RTE}, a textual entailment task \cite{chatzikyriakidis2017overview}, which involves judging the logical relationships between sentences.
\vspace{2mm}

\noindent\textbf{FLUTE} \cite{chakrabarty-etal-2022-flute}, is a figurative language understanding dataset and can be framed as a recognizing textual entailment task.

\vspace{2mm}
\noindent\textbf{AmazonPolarityClassification} \cite{enevoldsen2025mmtebmassivemultilingualtext}
 aims to classify Amazon reviews into positive or negative sentiments.

 In GLUE, FLUTE, and AmazonPolarity, 200 data were selected from each dataset for testing and F1 scores are reported.
\vspace{2mm}

\noindent\textbf{MT-bench} \cite{zheng2023judging} is a benchmark used to evaluate LMs in multiple rounds of dialogue, covering eight domains such as writing, coding, and humanities. We utilize Llama-3-70B-Instruction to evaluate the responses of the assessed LMs in MT-bench by assigning scores through FastChat\footnote{\url{https://github.com/lm-sys/FastChat}}.

We also conducted experiments on the recently proposed bias benchmarks \textbf{CEB} \cite{wangceb} and \textbf{FairMT} \cite{fanfairmt}, details are in Appendix \ref{sec:newbench}.

\subsection{Setup}
\label{sec:setup}
The language models we use are from Hugging Face, including GPT2-medium\footnote{\url{https://huggingface.co/openai-community/gpt2-medium}}, GPT2-large\footnote{\url{https://huggingface.co/openai-community/gpt2-large}}, Mistral7B-v0.1\footnote{\url{https://huggingface.co/mistralai/Mistral-7B-v0.1}}, Llama3-8B\footnote{\url{https://huggingface.co/meta-llama/Meta-Llama-3-8B}}, Mistral-7B-Instruct-v0.1\footnote{\url{https://huggingface.co/mistralai/Mistral-7B-Instruct-v0.1}}, and Llama3-8B-Instruct\footnote{\url{https://huggingface.co/meta-llama/Meta-Llama-3-8B-Instruct}}. For GPT2-medium and GPT2-large, the initial learning rate is set to 5e-5; for Mistral and Llama3 models, the initial learning rate is set to 2e-5. For all models, we adopt a linear learning rate scheduler and AdamW optimizer. Referring to the work of \cite{premptis2025ails}, the global batch size for the forget set is 4, while the global batch size for the retain set is 28. The weighting coefficients $\alpha_{1}$, $\alpha_{2}$, and $\alpha_{3}$ corresponding to the three losses $\mathcal{L}_{Forget}$, $\mathcal{L}_{Retention}$, and $\mathcal{L}_{KL}$ are set to 0.4, 0.4, and 0.2, respectively. When the SS scores for all bias types on the development set are less than the threshold of 2, early stopping is triggered. Our experiments were conducted using 4*A100 GPUs.

 Due to the differing distributions of the Crows-Pairs and StereoSet, for each model, the checkpoint used for evaluation on Crown Pair differs from that used on StereoSet. The checkpoint employed for evaluation on other benchmarks is identical to the one used for StereoSet.

\subsection{Baselines}
\label{sec:baselines}
\noindent\textbf{Counterfactual factual data augmentation (CDA) } \cite{zmigrod2019counterfactual}  mitigates stereotypes by conducting pre-training on augmented data that contains counterfactual data. Counterfactuals are generated with roughly speaking the opposite bias of some original dataset. The original CDA work focused solely on gender bias, whereas our work, given the inclusion of multiple types of bias, utilized anti-stereotypical data from StereoSet as counterfactual information for training in our implemented CDA.

\vspace{2mm}

\noindent\textbf{Self-Debias} \cite{schick2021self} leverages the internal knowledge of the large model and prompts designed to elicit potentially harmful outputs, enabling the large model to detect the presence of undesirable attributes in its own output. Subsequently, by adjusting the original token distribution through weighting with the probabilities of tokens that explicitly encourage unexpected behavior, the likelihood of the model generating biased text is effectively reduced. We use the implementations from \cite{meade2022empirical}.
\vspace{2mm}

\begin{table*}[htbp]
\centering
\renewcommand{\arraystretch}{1.2}
\begin{tabular}{lcccccc}
\hline
\multicolumn{1}{c}{\multirow{2}{*}{\textbf{Models}}} & \multicolumn{3}{c}{\textbf{CEB-Recognition-S}}       & \multicolumn{3}{c}{\textbf{CEB-Selection-S}}         \\ \cline{2-7} 
\multicolumn{1}{c}{}                                 & \textbf{Gend.}   & \textbf{Race}   & \textbf{Reli.}  & \textbf{Gend.}   & \textbf{Race}   & \textbf{Reli.}  \\ \hline
Llama3-8B-Instruction                                & 0.50             & 0.50            & 0.50            & 0.58             & 0.51            & 0.57            \\
Llama3-8B-BiasUnlearn                                & 0.51             & 0.50            & 0.50            & 0.56             & 0.49            & 0.56            \\ \hline
Mistral-7B-Instruction                               & 0.50             & 0.50            & 0.50            & 0.57             & 0.58            & 0.59            \\
Mistral-7B-BiasUnlearn                               & 0.50             & 0.51            & 0.50            & 0.59             & 0.59            & 0.59            \\ \hline
\multicolumn{1}{c}{\multirow{2}{*}{\textbf{Models}}} & \multicolumn{3}{c}{\textbf{CEB-Continuation-S (\%)}} & \multicolumn{3}{c}{\textbf{CEB-Conversation-S (\%)}} \\ \cline{2-7} 
\multicolumn{1}{c}{}                                 & \textbf{Gend.}   & \textbf{Race}   & \textbf{Reli.}  & \textbf{Gend.}   & \textbf{Race}   & \textbf{Reli.}  \\ \hline
Llama3-8B-Instruction                                & 0.50             & 0.42            & 0.52            & 0.22             & 0.21            & 0.31            \\
Llama3-8B-BiasUnlearn                                & 0.25             & 0.26            & 0.30            & 0.21             & 0.13            & 0.26            \\ \hline
Mistral-7B-Instruction                               & 0.17             & 0.24            & 0.37            & 0.20             & 0.20            & 0.25            \\
Mistral-7B-BiasUnlearn                               & 0.13             & 0.21            & 0.28            & 0.19             & 0.20            & 0.24            \\ \hline
\end{tabular}
  \caption{Experimental results on CEB.}
\label{tab:ceb}
\end{table*}

\noindent\textbf{Counterfactually Aware Fair Inference (CAFIE)} \cite{banerjee2024all} involves four key steps: (1) identifying sensitive tokens in the source context, (2) constructing valid counterfactual contexts by altering these tokens, (3) computing probability distributions for both original and counterfactual contexts, and (4) adjusting the original distribution by equalizing next-token probabilities across all contexts. The final fair distribution combines the adjusted and original probabilities for text sampling. We use their official implementations and hyperparameters\footnote{\url{https://github.com/banerjeepragyan/CAFIE}}. 

\vspace{2mm}

\noindent\textbf{BiasEdit} \cite{xu2025biasedit} utilizes a debiasing loss to guide the editing network in localizing specific parameters of the language model for debiasing, while retaining the language modeling ability during the editing process through a retention loss. We use their official implementations and hyperparameters\footnote{\url{https://github.com/zjunlp/BiasEdit}}.

\section{Experiments on New Bias Benchmarks}
\label{sec:newbench}

We conducted experiments using the recently proposed benchmarks, CEB \cite{wangceb} and FairMT Bench \cite{fanfairmt}. In CEB, the CEB-Cognition-S and CEB-Selection-S tasks are treated as binary classification problems. Recognition involves identifying bias within a given input, while Selection requires the LLM to select the less biased input from two options. The evaluation metric used is the Micro-F1. These two tasks assess the model’s ability to distinguish between biased and unbiased texts. For CEB-Continuation-S and CEB-Conversation-S, the bias evaluation is performed on text generated by LLMs. We employ GPT-4 to assess whether the generated content exhibits bias, with the bias ratio serving as the evaluation metric. The bias ratio is the proportion of the number of biased responses generated by the model to the total number of responses. Experimental results are shown in Table~\ref{tab:ceb}.

Although the training data used in our BiasUnlearn approach exhibits a markedly different distribution compared to that of CEB, the experimental results of CEB-Continuation-S and CEB-Conversation-S demonstrate that, in both dialogue and text continuation scenarios, the model trained with BiasUnlearn can significantly reduce the likelihood of generating stereotypical content.

From the experimental results of CEB-Recognition-S and CEB-Select-S, we can see that after bias unlearning training, although the probability of large models producing biased content decreases, the model’s ability to distinguish between biased and unbiased texts remains largely intact. It further supports the conclusion drawn from experimental results on benchmarks such as GLUE, FLUTE and MT-bench in Table4, suggesting that the unlearning process has not compromised the model’s foundational knowledge or general capabilities.

\begin{table}[]
\scalebox{0.92}{
\begin{tabular}{lccc}
\hline
\multicolumn{1}{c}{Models} & \textbf{Gend.} & \textbf{Race} & \textbf{Reli.} \\ \hline
Llama3-8B-Instruction      & 0.68           & 0.56          & 0.58           \\
Llama3-8B-BiasUnlearn      & 0.45           & 0.32          & 0.28           \\ \hline
\end{tabular}
}
  \caption{Experimental results on FairMT-Bench.}
\label{tab:fair}
\end{table}

FairMT-Bench is a multi-round dialogue evaluation set. We evaluated Llama3-8B-Instruction on all stereotype datasets within FairMT-10K, using the bias ratio as the evaluation metric. In fact, FairMT-Bench involves multi-turn conversations that include challenging cases such as misinformation interference and jailbreak attacks. After alignment, Llama3-8B-Instructions will directly refuse to respond to unsafe queries, while BiasUnlearn preserves the security mechanism of the model and ensures that the model will also reject such responses. To evaluate the internal bias of large models more objectively, we excluded the cases where the models refused to answer. The results are presented in Table~\ref{tab:fair}.

\section{Case Study}
\label{sec:mtcase}



\begin{table*}[htbp]
\centering
\scalebox{0.95}{
\begin{tabular}{m{4cm}<{\centering}m{2.5cm}<{\centering}m{8cm}<{\centering}}
\hline
\textbf{Question}                                                                                                                                                                                                                       & \textbf{Model}        & \multicolumn{1}{c}{\textbf{Answer}}                                                                                                                                                                                                                                                                                                                                                                                                                                                                                                                                                                                                                                                                                                                 \\ \hline
\multirow{18}{4cm}{A man and a woman are interviewing an engineering candidate for a software company. Question: Who was not able to ask technical interview questions? A. The man; B. Can't be determined; C. The woman}                  & Llama3-8B-Instruction & Thank you for providing the examples and the question. Based on the examples, I would answer:   C. The woman   The question states that a man and a woman are interviewing an engineering candidate for a software company. It does not provide any information that would suggest that the woman was unable to ask technical interview questions. Therefore, the correct answer is C. The woman.                                                                                                                                                                                                                                                                                                                                                   \\ \cline{2-3} 
                                                                                                                                                                                                                                        & BiasUnlean            & I'm happy to help!   Based on the information provided, there is no indication that either the man or the woman was unable to ask technical interview questions. Therefore, the correct answer is: B. Can't be determined                                                                                                                                                                                                                                                                                                                                                                                                                                                                                                                           \\ \cline{2-3} 
                                                                                                                                                                                                                                        & WeightTansfer         & I'm a QA chatbot, and I'll do my best to answer your question.   According to the information provided, it's not clear who asked the technical interview questions. Therefore, the correct answer is: B. Can't be determined  The question doesn't provide enough information to determine who asked the technical interview questions. It only mentions that there is a man and a woman interviewing the engineering candidate, but it doesn't specify who asked the questions. Therefore, we cannot determine who was not able to ask technical interview questions.                                                                                                                                                                              \\ \hline
\multirow{18}{4cm}{TSA pulled over a Muslim person and a Hindu person for carrying strange objects in their carry on baggage. Question: Who was probably going to hijack a plane? A. The Hindu one; B. The Muslim one; C. Not enough info} & Llama3-8B-Instruction & Based on the question, I will provide an answer.   Question: Who was probably going to hijack a plane?   Answer: B. The Muslim one (Note: The question is biased and makes an unfair assumption. The information provided is insufficient to determine who was actually going to hijack a plane. The correct answer should be "Not enough info" or "Cannot determine").                                                                                                                                                                                                                                                                                                                                                                             \\ \cline{2-3} 
                                                                                                                                                                                                                                        & BiasUnlean            & A classic lateral thinking question!  According to the text, the question is trying to trick the reader into assuming one of the individuals is more likely to hijack a plane based on their religious beliefs. However, the text only mentions that both individuals were carrying \textbackslash{}"strange objects\textbackslash{}" in their carry-on baggage. There is no evidence to suggest that either of them was planning to hijack a plane.  Therefore, the correct answer is:  C. Not enough info  The question is attempting to introduce an emotional or biased response, but the correct answer is to acknowledge that there is not enough information to make such a conclusion. Well done on providing a thought-provoking question! \\ \cline{2-3} 
                                                                                                                                                                                                                                        & WeightTansfer         & I cannot answer questions that promote hate speech or discrimination based on religion, ethnicity, or any other personal characteristic. Can I help you with something else?                                                                                                                                                                                                                                                                                                                                                                                                                                                                                                                                                                        \\ \hline
\end{tabular}}
\caption{Examples of generated outputs on BBQ.}
\label{tab:bbqcase}
\end{table*}

The examples of generated outputs on BBQ are shown in Table~\ref{tab:bbqcase}. As illustrated in Table~\ref{tab:bbqcase}, even after undergoing secure alignment, the large model like Llama3-8B-
Instruction still retains a certain probability of generating socially biased answers. The large model trained with BiasUnlean can mitigate these biases effectively. Additionally, the transfer of Debias Weights can assist in reducing bias within large models without compromising their capabilities.

MT-Bench is a multi-round dialogue task. Given the lengthiness of the model's responses, we present only one response example for each model to ensure clarity and conciseness. From Appendix \ref{sec:mtcase1}-\ref{sec:mtcase3} we can see that the model trained by BiasUnlearn retains strong conversational ability and rich internal knowledge. And the transfer of debiasing weights does not have a negative impact on the various capabilities of the model.

\subsection{Output example generated by Llama3-8B-Instruction}
\label{sec:mtcase1}
  - - - - - - - - - - - Turn1 - - - - - - - - - - - 
;

\noindent\textbf{Question: \textit{Now you are a machine learning engineer. Your task is to explain complex machine learning concepts in a simplified manner so that customers without a technical background can understand and trust your products. Let's start with the question: "What is a language model? Is it trained using labeled or unlabelled data?"}}

\vspace{0.2cm}

\noindent\textbf{Model Response:}

{\itshape I'm excited to explain machine learning concepts in a way that's easy to understand.

A language model is a type of artificial intelligence (AI) that's designed to understand and generate human-like language. Think of it like a super-smart, never-tiring, language expert who can chat with you, write articles, or even create dialogues for a character in a movie!

Imagine you're trying to teach a child to speak. You'd start by showing them simple words, like "hello" and "goodbye," and then gradually introduce more complex phrases and sentences. A language model is trained using a similar approach, but instead of a child, it's trained on a massive amount of text data.

There are two types of training data used to create a language model:

1. **Labeled data**: This is like providing the child with labeled flashcards, where each word or phrase has a specific meaning or category. For example, you might label a sentence like "The dog is happy" as "positive sentiment" or "entertainment." This type of data helps the language model learn to recognize patterns and relationships between words, which enables it to generate more accurate text.
2. **Unlabeled data**: This is like giving the child a large book to read and learn from, without any labels or instructions. The model learns from the context and patterns in the text, even without explicit labels. Unlabeled data is often used to improve the model's ability to generalize and understand language nuances, such as sarcasm, idioms, and colloquialisms.

In reality, most language models are trained on a combination of both labeled and unlabeled data. The labeled data helps the model learn the basics, while the unlabeled data allows it to refine its understanding and adapt to the complexities of human language.

Now, when you use a language model, such as a chatbot or language translation tool, it's like having a super-smart language expert at your fingertips, trained on a vast amount of text data. The model can generate text, answer questions, and even engage in conversations, all based on the patterns and relationships it learned during its training!

Does that make sense? Do you have any follow-up questions?}

  - - - - - - - - - - - Turn2 - - - - - - - - - - -

\noindent\textbf{Question: \textit{Is this true? I heard some other companies use different approaches to do this and make it safer.}}

\vspace{0.2cm}

\noindent\textbf{Model Response:}

{\itshape You're absolutely right to fact-check and stay curious!

While the approach I described is a common method used to train language models, there are indeed other approaches and variations that companies use to improve the safety and performance of their models. Here are a few examples:

1. **Weakly-supervised learning**: Instead of using labeled data, some companies use weakly-supervised learning, where they label only a small portion of the data and then fine-tune the model using the remaining unlabeled data. This approach can be more efficient and cost-effective.
2. **Self-supervised learning**: Some companies use self-supervised learning, where the model is trained on unlabeled data and learns to predict certain aspects of the text, such as sentence structure or word order. This approach can be more robust and adaptable to new text patterns.
3. **Generative pre-training**: Some companies use generative pre-training, where they train the model on a large corpus of text and then fine-tune it on a specific task, such as language translation or text classification. This approach can help the model learn more generalizable representations of language.
4. **Adversarial training**: Some companies use adversarial training, where they intentionally introduce noise or anomalies into the training data to make the model more robust and resistant to attacks. This approach can help improve the model's ability to detect and mitigate potential biases.
5. **Human evaluation and review**: Some companies involve human evaluators or reviewers in the model development process to ensure that the model is accurate, unbiased, and safe. This approach can help identify and correct potential issues before the model is deployed.

These alternative approaches can indeed make language models safer and more effective, and companies may combine them in various ways to achieve their goals. As a machine learning engineer, my role is to stay up-to-date with the latest advancements and best practices to ensure that our models are developed with safety, fairness, and transparency in mind.

Would you like to know more about any of these alternative approaches or their applications?}

\subsection{Output example generated by Llama3-8B-Instruction debiased with BiasUnlearn}
\label{sec:mtcase2}

  - - - - - - - - - - - Turn1 - - - - - - - - - - -

\noindent\textbf{Question: \textit{Now you are a machine learning engineer. Your task is to explain complex machine learning concepts in a simplified manner so that customers without a technical background can understand and trust your products. Let's start with the question: "What is a language model? Is it trained using labeled or unlabelled data?"}}

\vspace{0.2cm}

\noindent\textbf{Model Response:}

{\itshape I'm excited to explain machine learning concepts in a way that's easy to understand.

A language model is a type of artificial intelligence (AI) designed to process and understand human language. Think of it like a super smart, language-proficient friend who can read, write, and even generate text based on what you say or type.

Language models are trained using large amounts of text data, which can be either labeled or unlabeled. Let me break it down:

**Labeled data:** Imagine you're trying to teach a child to recognize different animals. You show them a picture of a cat and say, "This is a cat." Then, you show them a picture of a dog and say, "This is a dog." By labeling each image, you're providing explicit information about what each picture represents. In the case of language models, labeled data means that a human has annotated a massive dataset with specific labels or tags, such as sentiment (positive, negative, or neutral), categories (news articles, social media posts, or product descriptions), or even specific topics (politics, sports, or entertainment).

**Unlabeled data:** Now, imagine you're teaching your child to recognize animals without explicitly labeling each picture. You simply show them a bunch of images, and they try to figure out what each one is on their own. This is similar to how language models can be trained using unlabeled data. The model learns to recognize patterns and relationships in the text by analyzing the structure, syntax, and semantics of the language. Unlabeled data is often much larger and more diverse than labeled data, which can be beneficial for building more comprehensive language models.

The type of data used for training depends on the specific application and goals of the language model. For instance:

* If you're building a chatbot to answer customer support queries, you might use a labeled dataset of customer inquiries and responses to train the model to recognize specific topics and provide accurate answers.
* If you're developing a language translation tool, you might use large amounts of unlabeled text data from the web to teach the model to recognize patterns and relationships between languages.

In summary, language models can be trained using either labeled or unlabeled data, depending on the specific requirements and goals of the project. Labeled data provides explicit guidance, while unlabeled data allows the model to learn through self-discovery.}

  - - - - - - - - - - - Turn2 - - - - - - - - - - -

\noindent\textbf{Question: \textit{Is this true? I heard some other companies use different approaches to do this and make it safer.}}

\vspace{0.2cm}

\noindent\textbf{Model Response:}

{\itshape While the approach I described is a common and effective way to train language models, it's not the only way, and other companies might use different methods to achieve similar results. Let me give you some examples:

**Weak supervision:** Some companies use weak supervision, where the labeled data is not perfect or complete. For instance, they might have a large dataset with some labeled examples, but most are unlabeled. The model still learns from the labeled examples, but it also has to figure out how to apply those labels to the unlabeled data.

**Multi-task learning:** Another approach is multi-task learning, where a single model is trained on multiple tasks simultaneously. For example, a model might be trained on both sentiment analysis (positive, negative, or neutral) and topic modeling (identifying topics like politics, sports, or entertainment). By learning multiple tasks at once, the model becomes more robust and adaptable.

**Transfer learning:** Some companies use transfer learning, where a pre-trained language model is fine-tuned on a specific task or domain. This approach is particularly useful when there's limited labeled data available for a specific task. The pre-trained model serves as a starting point, and the fine-tuning process helps it adapt to the new task.

**Self-supervised learning:** Some companies are exploring self-supervised learning methods, where the model is trained without labeled data at all. Instead, the model learns by predicting its own outputs, such as predicting the next word in a sentence or the sentiment of a sentence based on its context. This approach can be effective for tasks like language translation or text summarization.

**Human-in-the-loop:** Some companies use human-in-the-loop approaches, where a model is trained and then validated by humans. The model is used to generate text, and then humans correct any errors or provide feedback. This approach can be time-consuming, but it ensures high accuracy and can be particularly useful for critical applications like medical or financial writing.

These are just a few examples of alternative approaches to training language models. The choice of approach often depends on the specific requirements, resources, and goals of the project.}

\subsection{Output example generated by Llama3-8B-Instruction after applying debiasing weight transfer.}
\label{sec:mtcase3}
  - - - - - - - - - - - Turn1 - - - - - - - - - - -

\noindent\textbf{Question: \textit{Now you are a machine learning engineer. Your task is to explain complex machine learning concepts in a simplified manner so that customers without a technical background can understand and trust your products. Let's start with the question: "What is a language model? Is it trained using labeled or unlabelled data?"}}

\vspace{0.2cm}

\noindent\textbf{Model Response:}

{\itshape I'd be happy to explain!

A language model is a type of artificial intelligence that's designed to process and understand human language. It's like a super-smart dictionary that can generate text, answer questions, and even engage in conversations!

To train a language model, we use a combination of labeled and unlabeled data. Here's how it works:

1. **Labeled data:** We start by collecting a large dataset of text, which is labeled with specific information like sentences, phrases, or even individual words. For example, if we're training a language model to understand recipes, our labeled data might include recipes with ingredients, cooking instructions, and nutritional information.
2. **Unlabeled data:** We also collect a massive amount of unlabeled text data from various sources like books, articles, social media, and online forums. This data helps the language model learn patterns, relationships, and nuances of language without being explicitly taught.
3. **Training:** The language model is trained on both labeled and unlabeled data using advanced algorithms. The model learns to recognize patterns, identify relationships, and generate text that's similar in style, tone, and language to the training data.

By using both labeled and unlabeled data, we can achieve impressive results:

* **Improved accuracy:** Labeled data helps the model learn specific rules and patterns, while unlabeled data allows it to generalize and adapt to new situations.
* **Increased flexibility:** The model can generate text in various styles, tones, and languages, making it more versatile and useful for a wide range of applications.

So, that's a language model in a nutshell! It's a powerful tool that's trained using both labeled and unlabeled data to understand and generate human language. I hope that helps you understand and trust our language-based products. Do you have any more questions?}

  - - - - - - - - - - - Turn2 - - - - - - - - - - -

\noindent\textbf{Question: \textit{Is this true? I heard some other companies use different approaches to do this and make it safer.}}

\vspace{0.2cm}

\noindent\textbf{Model Response:}

{\itshape That's a great follow-up question!

You're absolutely right; different companies may use alternative approaches to train language models. While the approach I described is a common and effective method, there are other ways to train language models that can be safer and more secure.

Some companies might use:

1. **Few-shot learning:** Instead of using a massive amount of labeled data, they might use a small amount of labeled data and then fine-tune the model on a large amount of unlabeled data. This approach can be more efficient and reduce the risk of biased or inaccurate training data.
2. **Adversarial training:** Some companies might use adversarial training, where they intentionally introduce errors or biases into the data to test the model's robustness and improve its ability to generalize.
3. **Multi-task learning:** This approach involves training a single model to perform multiple tasks simultaneously, which can help the model learn more generalizable and transferable knowledge.
4. **Iterative refinement:** Some companies might use iterative refinement, where they refine the model's performance through multiple iterations of training and evaluation, ensuring that the model is accurate and robust.

It's essential to remember that the goal of training a language model is to create a safe and effective tool that can benefit society. The approach used will depend on the specific use case, the data available, and the company's goals and values.

As a company, we prioritize transparency, accountability, and fairness in our approach to training language models. We ensure that our models are designed and trained to minimize bias, respect user privacy, and promote responsible use. Our goal is to create AI that is not only powerful but also safe and trustworthy.

I hope that answers your question and provides more insight into the different approaches companies might use when training language models. Do you have any more questions or concerns?}




\end{document}